\documentclass{article}

\usepackage{times}
\usepackage{graphicx}
\usepackage{amsmath}
\usepackage{amssymb}
\usepackage{subfig}
\usepackage{caption}
\usepackage{gensymb}

\usepackage{natbib}

\usepackage{algorithm}
\usepackage{algorithmic}

\usepackage{hyperref}

\usepackage[accepted]{icml2016_arxiv}

\icmltitlerunning{One-shot learning with Memory-Augmented Neural Networks}

\begin{document} 

\twocolumn[
\icmltitle{One-shot Learning with Memory-Augmented Neural Networks}

\icmlauthor{Adam Santoro}{adamsantoro@google.com}
\icmladdress{Google DeepMind}
\icmlauthor{Sergey Bartunov}{sbos@sbos.in}
\icmladdress{Google DeepMind, National Research University Higher School of Economics (HSE)}
\icmlauthor{Matthew Botvinick}{botvinick@google.com}
\icmlauthor{Daan Wierstra}{wierstra@google.com}
\icmlauthor{Timothy Lillicrap}{countzero@google.com}
\icmladdress{Google DeepMind}

\icmlkeywords{meta-learning, neural turing machine, memory}

\vskip 0.3in
]

\begin{abstract} 
Despite recent breakthroughs in the applications of deep neural networks, one setting that presents a persistent challenge is that of ``one-shot learning." Traditional gradient-based networks require a lot of data to learn, often through extensive iterative training. When new data is encountered, the models must inefficiently relearn their parameters to adequately incorporate the new information without catastrophic interference. Architectures with augmented memory capacities, such as Neural Turing Machines (NTMs), offer the ability to quickly encode and retrieve new information, and hence can potentially obviate the downsides of conventional models. Here, we demonstrate the ability of a memory-augmented neural network to rapidly assimilate new data, and leverage this data to make accurate predictions after only a few samples. We also introduce a new method for accessing an external memory that focuses on memory content, unlike previous methods that additionally use memory location-based focusing mechanisms. 

\end{abstract} 

\section{Introduction}
\label{introduction}
The current success of deep learning hinges on the ability to apply gradient-based optimization to high-capacity models. This approach has achieved impressive results on many large-scale supervised tasks with raw sensory input, such as image classification \cite{he2015delving}, speech recognition \cite{yu2012automatic}, and games \cite{mnih2015human,silver2016mastering}. Notably, performance in such tasks is typically evaluated after extensive, incremental training on large data sets. In contrast, many problems of interest require rapid inference from small quantities of data. In the limit of ``one-shot learning," single observations should result in abrupt shifts in behavior.

This kind of flexible adaptation is a celebrated aspect of human learning \cite{jankowski2011meta}, manifesting in settings ranging from motor control \cite{braun2009motor} to the acquisition of abstract concepts \cite{lake2015human}. Generating novel behavior based on inference from a few scraps of information -- e.g., inferring the full range of applicability for a new word, heard in only one or two contexts -- is something that has remained stubbornly beyond the reach of contemporary machine intelligence. It appears to present a particularly daunting challenge for deep learning. In situations when only a few training examples are presented one-by-one, a straightforward gradient-based solution is to completely re-learn the parameters from the data available at the moment. Such a strategy is prone to poor learning, and/or catastrophic interference. In view of these hazards, non-parametric methods are often considered to be better suited.

However, previous work does suggest one potential strategy for attaining rapid learning from sparse data, and hinges on the notion of \textit{meta-learning} \cite{thrun1998lifelong,vilalta2002perspective}. Although the term has been used in numerous senses \cite{schmidhuber1997shifting,caruana1997multitask,schweighofer2003meta,brazdil2003ranking}, meta-learning generally refers to a scenario in which an agent learns at two levels, each associated with different time scales. Rapid learning occurs \textit{within} a task, for example, when learning to accurately classify within a particular dataset. This learning is guided by knowledge accrued more gradually \textit{across} tasks, which captures the way in which task structure varies across target domains \cite{giraud2004introduction,rendell1987layered,thrun1998lifelong}. Given its two-tiered organization, this form of meta-learning is often described as ``learning to learn.''

It has been proposed that neural networks with memory capacities could prove quite capable of meta-learning \cite{hochreiter2001learning}. These networks shift their bias through weight updates, but also modulate their output by learning to rapidly cache representations in memory stores \cite{hochreiter1997long}. For example, LSTMs trained to meta-learn can quickly learn never-before-seen quadratic functions with a low number of data samples \cite{hochreiter2001learning}. 

Neural networks with a memory capacity provide a promising approach to meta-learning in deep networks. However, the specific strategy of using the memory inherent in unstructured recurrent architectures is unlikely to extend to settings where each new task requires significant amounts of new information to be rapidly encoded. A scalable solution has a few necessary requirements: First, information must be stored in memory in a representation that is both stable (so that it can be reliably accessed when needed) and element-wise addressable (so that relevant pieces of information can be accessed selectively). Second, the number of parameters should not be tied to the size of the memory. These two characteristics do not arise naturally within standard memory architectures, such as LSTMs. However, recent architectures, such as Neural Turing Machines (NTMs) \cite{graves2014neural} and memory networks \cite{weston2014memory}, meet the requisite criteria. And so, in this paper we revisit the meta-learning problem and setup from the perspective of a highly capable memory-augmented neural network (MANN) (note: here on, the term MANN will refer to the class of external-memory equipped networks, and not other ``internal" memory-based architectures, such as LSTMs). 

We demonstrate that MANNs are capable of meta-learning in tasks that carry significant short- and long-term memory demands. This manifests as successful classification of never-before-seen Omniglot classes at human-like accuracy after only a few presentations, and principled function estimation based on a small number of samples. Additionally, we outline a memory access module that emphasizes memory access by content, and not additionally on memory location, as in original implementations of the NTM \cite{graves2014neural}. Our approach combines the best of two worlds: the ability to slowly learn an abstract method for obtaining useful representations of raw data, via gradient descent, and the ability to rapidly bind never-before-seen information after a single presentation, via an external memory module. The combination supports robust meta-learning, extending the range of problems to which deep learning can be effectively applied.

\section{Meta-Learning Task Methodology}
\label{sec:meta-learning-task}
\begin{figure*}[ht]
\begin{center}
\subfloat[][Task setup]{\includegraphics[width=1.1\columnwidth]{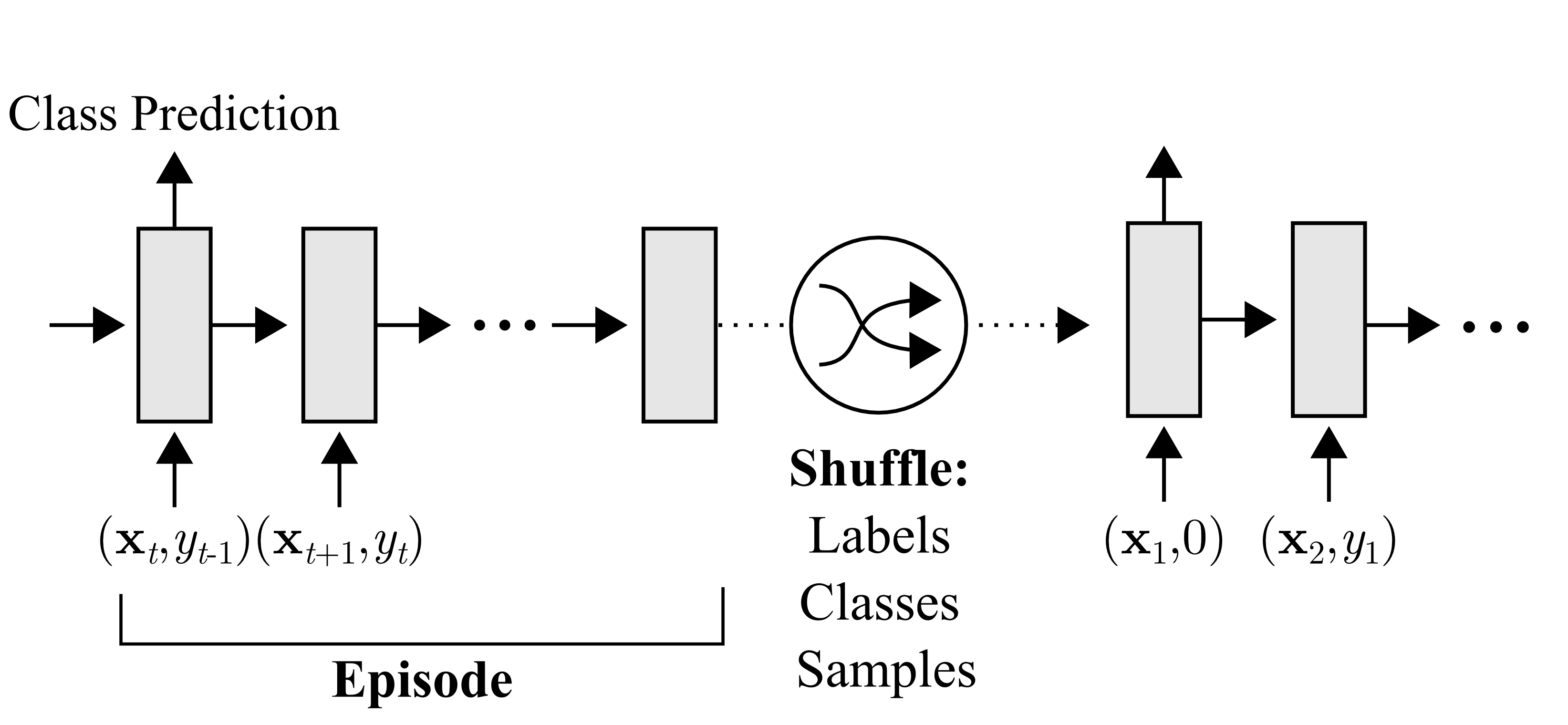}} \qquad
\subfloat[][Network strategy]{\includegraphics[width=0.89\columnwidth]{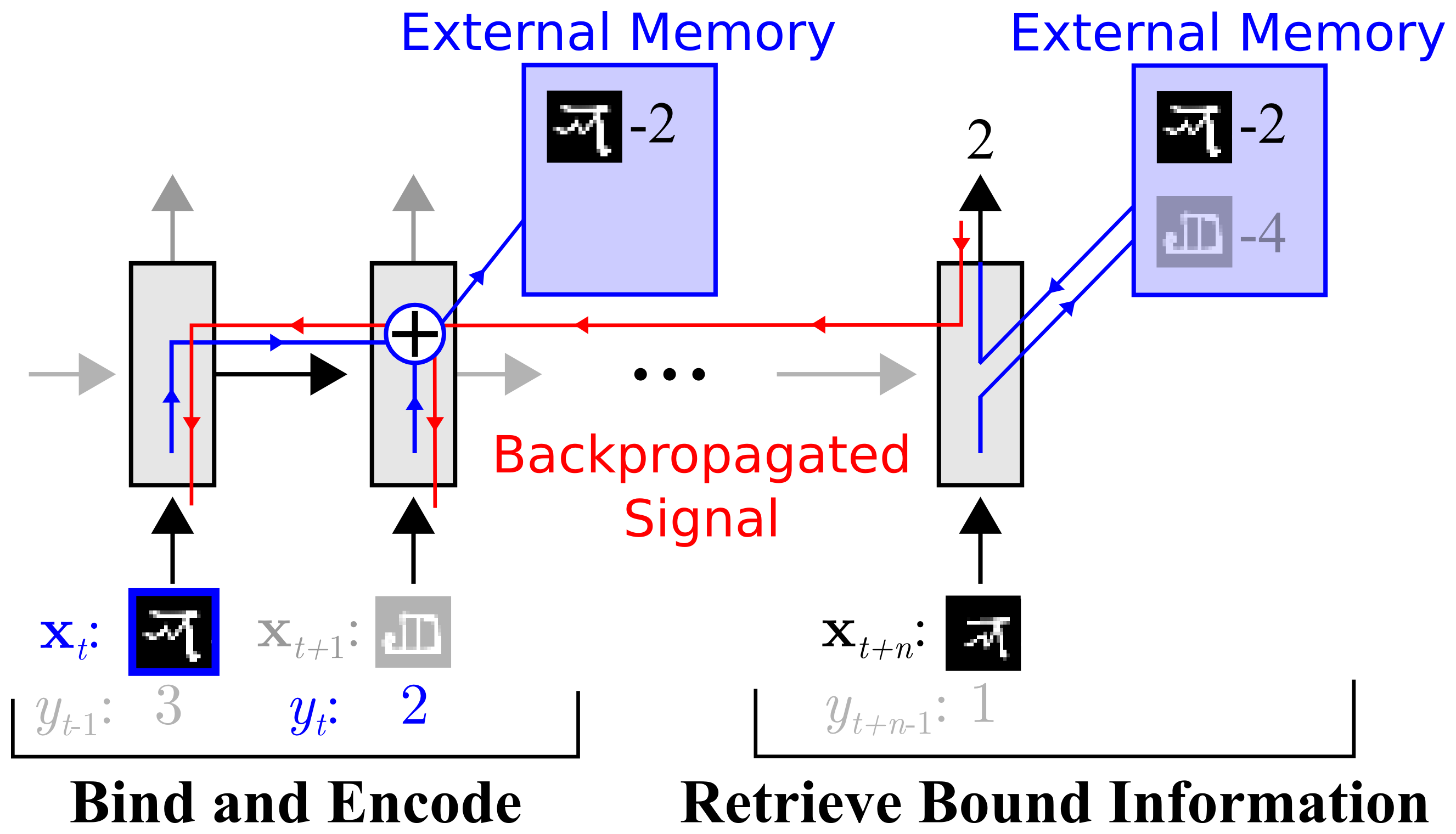}}
\caption{Task structure. (a) Omniglot images (or $x$-values for regression), $\mathbf{x}_t$, are presented with time-offset labels (or function values), $y_{t-1}$, to prevent the network from simply mapping the class labels to the output. From episode to episode, the classes to be presented in the episode, their associated labels, and the specific samples are all shuffled. (b) A successful strategy would involve the use of an external memory to store bound sample representation-class label information, which can then be retrieved at a later point for successful classification when a sample from an already-seen class is presented. Specifically, sample data $x_t$ from a particular time step should be bound to the appropriate class label $y_t$, which is presented in the subsequent time step. Later, when a sample from this same class is seen, it should retrieve this bound information from the external memory to make a prediction. Backpropagated error signals from this prediction step will then shape the weight updates from the earlier steps in order to promote this binding strategy. }
\label{fig:task}
\end{center}
\vskip -0.2in
\end{figure*}

Usually, we try to choose parameters $\theta$ to minimize a learning cost $\mathcal{L}$ across some dataset $D$. However, for meta-learning, we choose parameters to reduce the expected learning cost across a distribution of datasets $p(D)$:
\begin{align}
    \theta^* = \text{argmin}_\theta E_{D \sim p(D)} [\mathcal{L}(D; \theta)].
\end{align}

To accomplish this, proper task setup is critical \cite{hochreiter2001learning}. In our setup, a task, or episode, involves the presentation of some dataset $D=\{d_t\}_{t=1}^T=\{(\mathbf{x}_t, y_t)\}_{t=1}^T$. For classification, $y_t$ is the class label for an image $\mathbf{x}_t$, and for regression, $y_t$ is the value of a hidden function for a vector with real-valued elements $\mathbf{x}_t$, or simply a real-valued number $x_t$ (here on, for consistency, $\mathbf{x}_t$ will be used). In this setup, $y_t$ is both a target, and is presented as input along with $\mathbf{x}_t$, in a temporally offset manner; that is, the network sees the input sequence $(\mathbf{x}_1, \text{null}), (\mathbf{x}_2, y_1), \ldots, (\mathbf{x}_T, y_{T-1})$. And so, at time $t$ the correct label for the previous data sample ($y_{t-1}$) is provided as input along with a new query $\mathbf{x}_t$ (see Figure \ref{fig:task} (a)). The network is tasked to output the appropriate label for $\mathbf{x}_t$ (i.e., $y_t$) at the given timestep. Importantly, labels are shuffled from dataset-to-dataset. This prevents the network from slowly learning sample-class bindings in its weights. Instead, it must learn to hold data samples in memory until the appropriate labels are presented at the next time-step, after which sample-class information can be bound and stored for later use (see Figure \ref{fig:task} (b)). Thus, for a given episode, ideal performance involves a random guess for the first presentation of a class (since the appropriate label can not be inferred from previous episodes, due to label shuffling), and the use of memory to achieve perfect accuracy thereafter. Ultimately, the system aims at modelling the predictive distribution $p(y_t | \mathbf{x}_t, D_{1:t-1}; \theta)$, inducing a corresponding loss at each time step. 

This task structure incorporates exploitable meta-knowledge: a model that meta-learns would learn to bind data representations to their appropriate labels regardless of the actual content of the data representation or label, and would employ a general scheme to map these bound representations to appropriate classes or function values for prediction.

\section{Memory-Augmented Model}
\subsection{Neural Turing Machines}
The Neural Turing Machine is a fully differentiable implementation of a MANN. It consists of a controller, such as a feed-forward network or LSTM, which interacts with an external memory module using a number of read and write heads \cite{graves2014neural}. Memory encoding and retrieval in a NTM external memory module is rapid, with vector representations being placed into or taken out of memory potentially every time-step. This ability makes the NTM a perfect candidate for meta-learning and low-shot prediction, as it is capable of both long-term storage via slow updates of its weights, and short-term storage via its external memory module. Thus, if a NTM can learn a general strategy for the types of representations it should place into memory and how it should later use these representations for predictions, then it may be able use its speed to make accurate predictions of data that it has only seen once. 

The controllers employed in our model are are either LSTMs, or feed-forward networks. The controller interacts with an external memory module using read and write heads, which act to retrieve representations from memory or place them into memory, respectively. Given some input, $\mathbf{x}_t$, the controller produces a key, $\mathbf{k}_t$, which is then either stored in a row of a memory matrix $\mathbf{M}_t$, or used to retrieve a particular memory, $i$, from a row; i.e., $\mathbf{M}_t(i)$. When retrieving a memory, $\mathbf{M}_t$ is addressed using the cosine similarity measure,
\begin{align}
    K\big(\mathbf{k}_t, \mathbf{M}_{t}(i)\big) = \frac{\mathbf{k}_t \cdot \mathbf{M}_{t}(i)}{\parallel\mathbf{k}_t\parallel\parallel\mathbf{M}_{t}(i)\parallel},
\end{align}

which is used to produce a read-weight vector, $\mathbf{w}_t^r$, with elements computed according to a softmax:
\begin{align}
    w_t^{r}(i) \leftarrow \frac{\text{exp}\big(K\big(\mathbf{k}_t, \mathbf{M}_{t}(i)\big)\big)}{\sum_j \text{exp} \big( K\big(\mathbf{k}_t, \mathbf{M}_{t}(j)\big)\big)}.
\end{align}

A memory, $\mathbf{r}_t$, is retrieved using this weight vector:
\begin{align}
    \mathbf{r}_t \leftarrow \sum_i {w}_t^r(i)\mathbf{M}_{t}(i).
\end{align}
This memory is used by the controller as the input to a classifier, such as a softmax output layer, and as an additional input for the next controller state.

\subsection{Least Recently Used Access}
\label{sec:LRUA}

In previous instantiations of the NTM \cite{graves2014neural}, memories were addressed by both content and location. Location-based addressing was used to promote iterative steps, akin to running along a tape, as well as long-distance jumps across memory. This method was advantageous for sequence-based prediction tasks. However, this type of access is not optimal for tasks that emphasize a conjunctive coding of information independent of sequence. As such, writing to memory in our model involves the use of a newly designed access module called the Least Recently Used Access (LRUA) module. 

The LRUA module is a pure content-based memory writer that writes memories to either the least used memory location or the most recently used memory location. This module emphasizes accurate encoding of relevant (i.e., recent) information, and pure content-based retrieval. New information is written into rarely-used locations, preserving recently encoded information, or it is written to the last used location, which can function as an update of the memory with newer, possibly more relevant information. The distinction between these two options is accomplished with an interpolation between the previous read weights and weights scaled according to usage weights $\mathbf{w}_t^u$. These usage weights are updated at each time-step by decaying the previous usage weights and adding the current read and write weights:
\begin{align}
    \mathbf{w}_t^u \leftarrow \gamma \mathbf{w}_{t-1}^u + \mathbf{w}_t^r + \mathbf{w}_t^w.
\end{align}

Here, $\gamma$ is a decay parameter and $\mathbf{w}_t^r$ is computed as in (3). The \textit{least-used} weights, $\mathbf{w}_t^{lu}$, for a given time-step can then be computed using $\mathbf{w}_{t}^u$. First, we introduce the notation $m(\mathbf{v}, n)$ to denote the $n^{th}$ smallest element of the vector $\mathbf{v}$. Elements of $\mathbf{w}_{t}^{lu}$ are set accordingly:
\begin{align}
    w_{t}^{lu}(i) = 
    \left\{
	\begin{array}{ll}
		0  & \mbox{if } w_{t}^{u}(i) > m(\mathbf{w}_{t}^{u}, n) \\
		1  & \mbox{if } w_{t}^{u}(i) \le m(\mathbf{w}_{t}^{u}, n)
	\end{array}
\right. ,
\end{align}
where $n$ is set to equal the number of reads to memory. To obtain the write weights $\mathbf{w}_t^w$, a learnable sigmoid gate parameter is used to compute a convex combination of the previous read weights and previous least-used weights:
\begin{align}
    \mathbf{w}_t^w \leftarrow \sigma(\alpha) \mathbf{w}_{t-1}^r + (1-\sigma(\alpha))\mathbf{w}_{t-1}^{lu}.
\end{align}

Here, $\sigma(\cdot)$ is a sigmoid function, $\frac{1}{1+e^{-x}}$, and $\alpha$ is a scalar gate parameter to interpolate between the weights. Prior to writing to memory, the least used memory location is computed from $\mathbf{w}_{t-1}^u$ and is set to zero. Writing to memory then occurs in accordance with the computed vector of write weights:
\begin{align}
    \mathbf{M}_t(i) \leftarrow \mathbf{M}_{t-1}(i) + w_t^{w}(i) \mathbf{k}_{t}, \forall i
\end{align}

Thus, memories can be written into the zeroed memory slot or the previously used slot; if it is the latter, then the least used memories simply get erased.

\section{Experimental Results}

\subsection{Data}
Two sources of data were used: Omniglot, for classification, and sampled functions from a Gaussian process (GP) with fixed hyperparameters, for regression. The Omniglot dataset consists of over 1600 separate classes with only a few examples per class, aptly lending to it being called the transpose of MNIST \cite{lake2015human}. To reduce the risk of overfitting, we performed data augmentation by randomly translating and rotating character images. We also created new classes through $90^{\circ}$, $180^{\circ}$ and $270^{\circ}$ rotations of existing data. The training of all models was performed on the data of 1200 original classes (plus augmentations), with the rest of the 423 classes (plus augmentations) being used for test experiments. In order to reduce the computational time of our experiments we downscaled the images to $20\times20$.

\begin{figure*}[t]
\centering
\subfloat[][LSTM, five random classes/episode, one-hot vector labels]{\includegraphics[width=0.99\columnwidth]{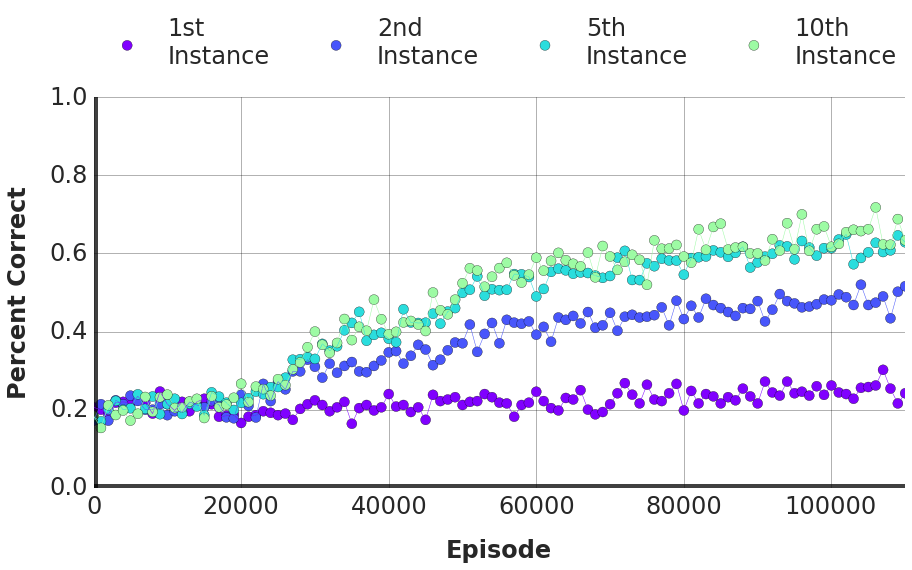}} \qquad
\subfloat[][MANN, five random classes/episode, one-hot vector labels]{\includegraphics[width=0.99\columnwidth]{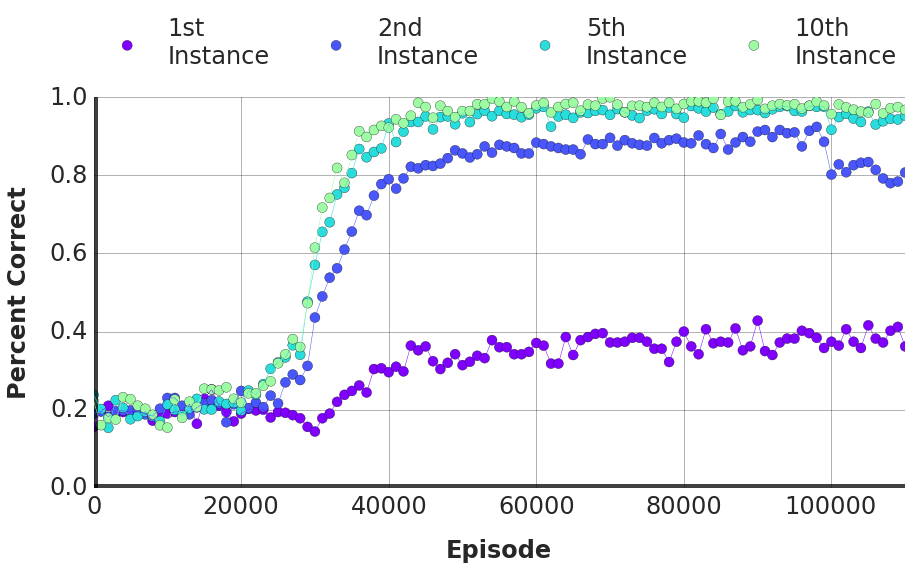}}
\newline
\subfloat[][LSTM, fifteen classes/episode, five-character string labels]{\includegraphics[width=0.99\columnwidth]{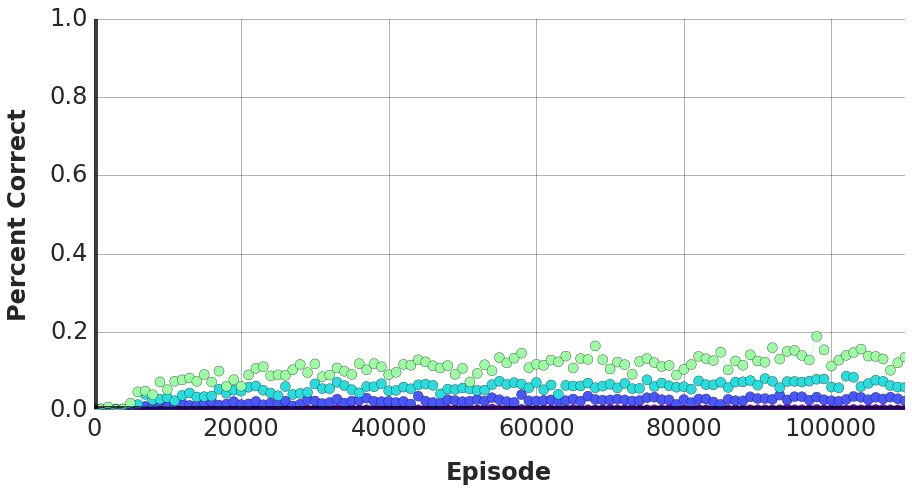}}\qquad
\subfloat[][MANN, fifteen classes/episode, five-character string labels]{\includegraphics[width=0.99\columnwidth]{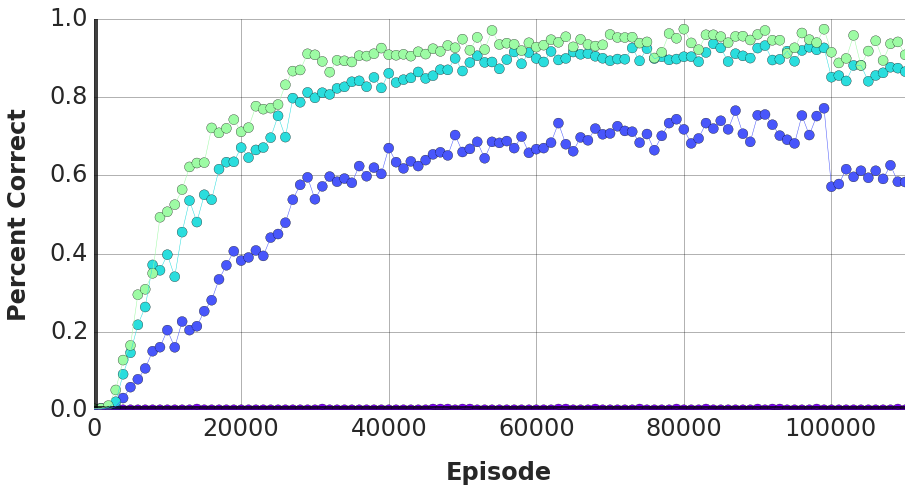}}
\caption{Omniglot classification. The network was given either five (a-b) or up to fifteen (c-d) random classes per episode, which were of length 50 or 100 respectively. Labels were one-hot vectors in (a-b), and five-character strings in (c-d). In (b), first instance accuracy is above chance, indicating that the MANN is performing ``educated guesses'' for new classes based on the classes it has already seen and stored in memory. In (c-d), first instance accuracy is poor, as is expected, since it must make a guess from 3125 random strings. Second instance accuracy, however, approaches 80\% during training for the MANN (d). At the 100,000 episode mark the network was tested, without further learning, on distinct classes withheld from the training set, and exhibited comparable performance.}%
\label{fig:omniglot-classification}%
\end{figure*}

\subsection{Omniglot Classification}
We performed a number of iterations of the basic task described in Section \ref{sec:meta-learning-task}. First, our MANN was trained using one-hot vector representations as class labels (Figure \ref{fig:omniglot-classification}). After training on 100,000 episodes with five randomly chosen classes with randomly chosen labels, the network was given a series of test episodes. In these episodes, no further learning occurred, and the network was to predict the class labels for never-before-seen classes pulled from a disjoint test set from within Omniglot. The network exhibited high classification accuracy on just the second presentation of a sample from a class within an episode (82.8\%), reaching up to 94.9\% accuracy by the fifth instance and 98.1\% accuracy by the tenth. 

For classification using one-hot vector representations, one relevant baseline is human performance. Participants were first given instructions detailing the task: an image would appear, and they must choose an appropriate digit label from the integers 1 through 5. Next, the image was presented and they were to make an un-timed prediction as to its class label. The image then disappeared, and they were given visual feedback as to their correctness, along with the correct label. The correct label was presented regardless of the accuracy of their prediction, allowing them to further reinforce correct decisions. After a short delay of two seconds, a new image appeared and they repeated the prediction process. The participants were not permitted to view previous images, or to use a scratch pad for externalization of memory. Performance of the MANN surpassed that of a human on each instance. Interestingly, the MANN displayed better than random guessing on the first instance within a class. Seemingly, it employed a strategy of educated guessing; if a particular sample produced a key that was a poor match to any of the bindings stored in external memory, then the network was less likely to choose the class labels associated with these stored bindings, and hence increased its probability of correctly guessing this new class on the first instance. A similar strategy was reported qualitatively by the human participants. We were unable to accumulate an appreciable amount of data from participants on the fifteen class case, as it proved much too difficult and highly demotivating. For all intents and purposes, as the number of classes scale to fifteen and beyond, this type of binding surpasses human working memory capacity, which is limited to storing only a handful of arbitrary bindings \cite{cowan2010magical}. 

\begin{table}[t]
\caption{Test-set classification accuracies for humans compared to machine algorithms trained on the Omniglot dataset, using one-hot encodings of labels and five classes presented per episode.}
\label{omniglot-test--1hot-table}
\vskip 0.15in
\begin{center}
\begin{small}
\begin{sc}
\begin{tabular}{l|p{5mm}p{5mm}p{5mm}p{5mm}p{5mm}p{5mm}}
\hline
& \multicolumn{6}{c}{Instance (\% Correct)} \\
Model & 1\textsuperscript{st} & 2\textsuperscript{nd} & 3\textsuperscript{rd} & 4\textsuperscript{th} & 5\textsuperscript{th} & 10\textsuperscript{th} \\
\hline
\abovespace
Human       & 34.5  & 57.3  & 70.1  & 71.8  & 81.4  & 92.4  \\
Feedforward & 24.4  & 19.6  & 21.1  & 19.9  & 22.8  & 19.5  \\
LSTM        & 24.4  & 49.5  & 55.3  & 61.0  & 63.6  & 62.5  \\
MANN         & \textbf{36.4}  & \textbf{82.8}  & \textbf{91.0}  & \textbf{92.6}  & \textbf{94.9}  & \textbf{98.1}  \\
\hline
\end{tabular}
\end{sc}
\end{small}
\end{center}
\vskip -0.1in
\end{table}


Since learning the weights of a classifier using large one-hot vectors becomes increasingly difficult with scale, a different approach for labeling classes was employed so that the number of classes presented in a given episode could be arbitrarily increased. These new labels consisted of strings of five characters, with each character assuming one of five possible values. Characters for each label were uniformly sampled from the set \{`a', `b', `c', `d', `e'\}, producing random strings such as `ecdba'. Strings were represented as concatenated one-hot vectors, and hence were of length 25 with five elements assuming a value of 1, and the rest 0. This combinatorial approach allows for 3125 possible labels, which is nearly twice the number of classes in the dataset. Therefore, the probability that a given class assumed the same label in any two episodes throughout training was greatly reduced. This also meant, however, that the guessing strategy exhibited by the network for the first instance of a particular class within an episode would probably be abolished. Nonetheless, this method allowed for episodes containing a large number of unique classes.

To confirm that the network was able to learn using these class representations, the previously described experiment was repeated (See Table \ref{omniglot-string-table}). Notably, a MANN with a standard NTM access module was unable to reach comparable performance to a MANN with LRU Access. Given this success, the experiment was scaled to up to fifteen unique classes presented in episodes of length 100, with the network exhibiting similar performance. 

We considered a set of baselines, such as a feed-forward RNN, LSTM, and a nonparametric nearest neighbours classifier that used either raw-pixel input or features extracted by an autoencoder. The autoencoder consisted of an encoder and decoder each with two 200-unit layers with leaky ReLU activations, and an output bottleneck layer of 32 units. The resultant architecture contained significantly more parameters than the MANN and, additionally, was allowed to train on three times as much augmented data. The highest accuracies from our experiments are reported, which were achieved using a single nearest neighbour for prediction and features from the output bottleneck layer of the autoencoder. Importantly, the nearest neighbour classifier had an unlimited amount of memory, and could automatically store and retrieve all previously seen examples. This provided the kNN with an distinct advantage, even when raw pixels were used as input representations. Although using rich features extracted by the autoencoder further improved performance, the kNN baseline was clearly outperformed by the MANN.

\subsubsection{Persistent Memory Interference}
A good strategy to employ in this classification task, and the strategy that was artificially imposed thus-far, is to wipe the external memory from episode to episode. Since each episode contains unique classes, with unique labels, any information persisting in memory across episodes inevitably acts as interference for the episode at hand. To test the effects of memory interference, we performed the classification task without wiping the external memory between episodes. 

\begin{figure}[ht]
\centering
\subfloat[][Five classes per episode]{\includegraphics[width=0.99\columnwidth]{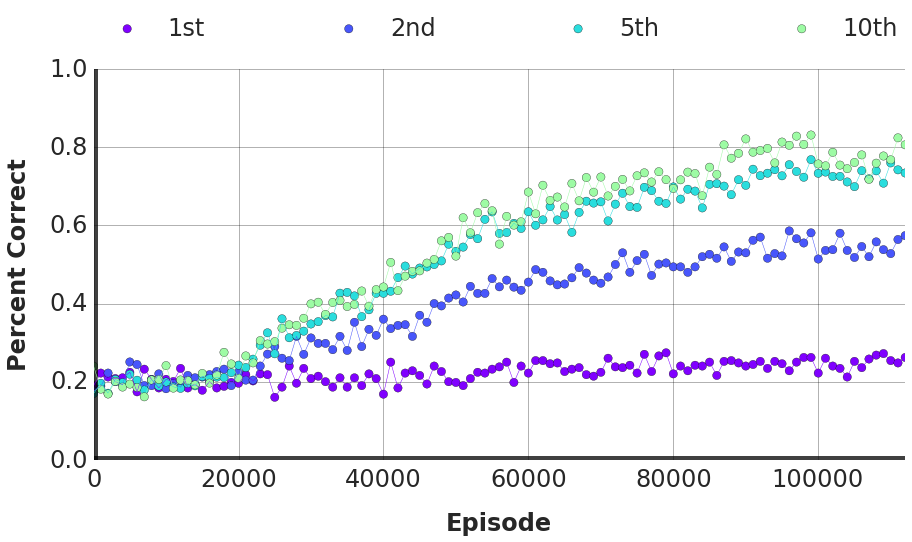}}
\qquad
\subfloat[][Ten classes per episode]{\includegraphics[width=0.99\columnwidth]{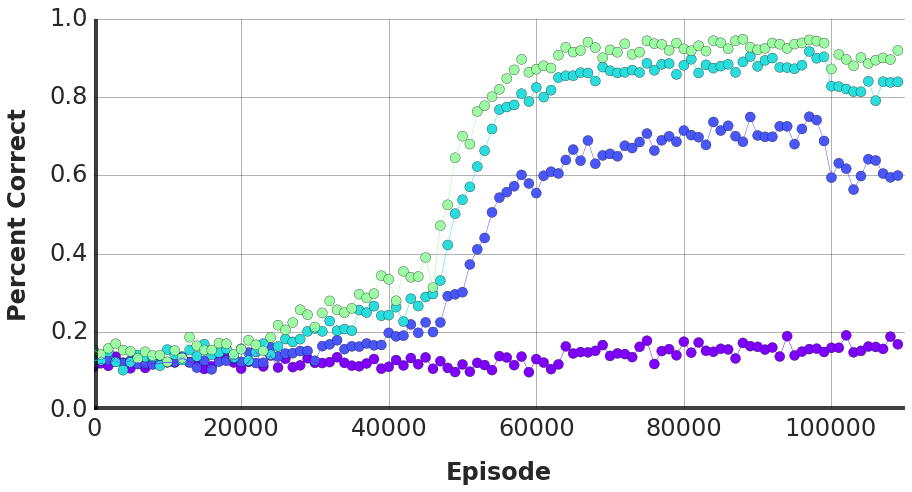}}

\caption{Persistent memory. To test the effects of interference, we did not wipe the external memory store from episode-to-episode. The network indeed struggled in this task (a), but nonetheless was able to perform comparably under certain setups, such as when episodes included ten classes and were of length 75 (b).}%
\label{fig:omniglot-persistent}%
\end{figure}

This task proved predictably difficult, and the network was less robust in its ability to achieve accurate classification (Figure \ref{fig:omniglot-persistent}). For example, in the case of learning one-hot vector labels in an episode that contained five unique classes, learning progressed much slower than in the memory-wipe condition, and did not produce the characteristic fast spike in accuracy seen in the memory-wipe condition (Figure \ref{fig:omniglot-classification}). Interestingly, there were conditions in which learning was not compromised appreciably. In the case of learning ten unique classes in episodes of length 75, for example, classification accuracy reached comparable levels. Exploring the requirements for robust performance is a topic of future work.

\subsubsection{Curriculum Training}
\begin{figure}[ht]
\centering
\subfloat[][One additional class per 10,000 episodes]{\includegraphics[width=0.98\columnwidth]{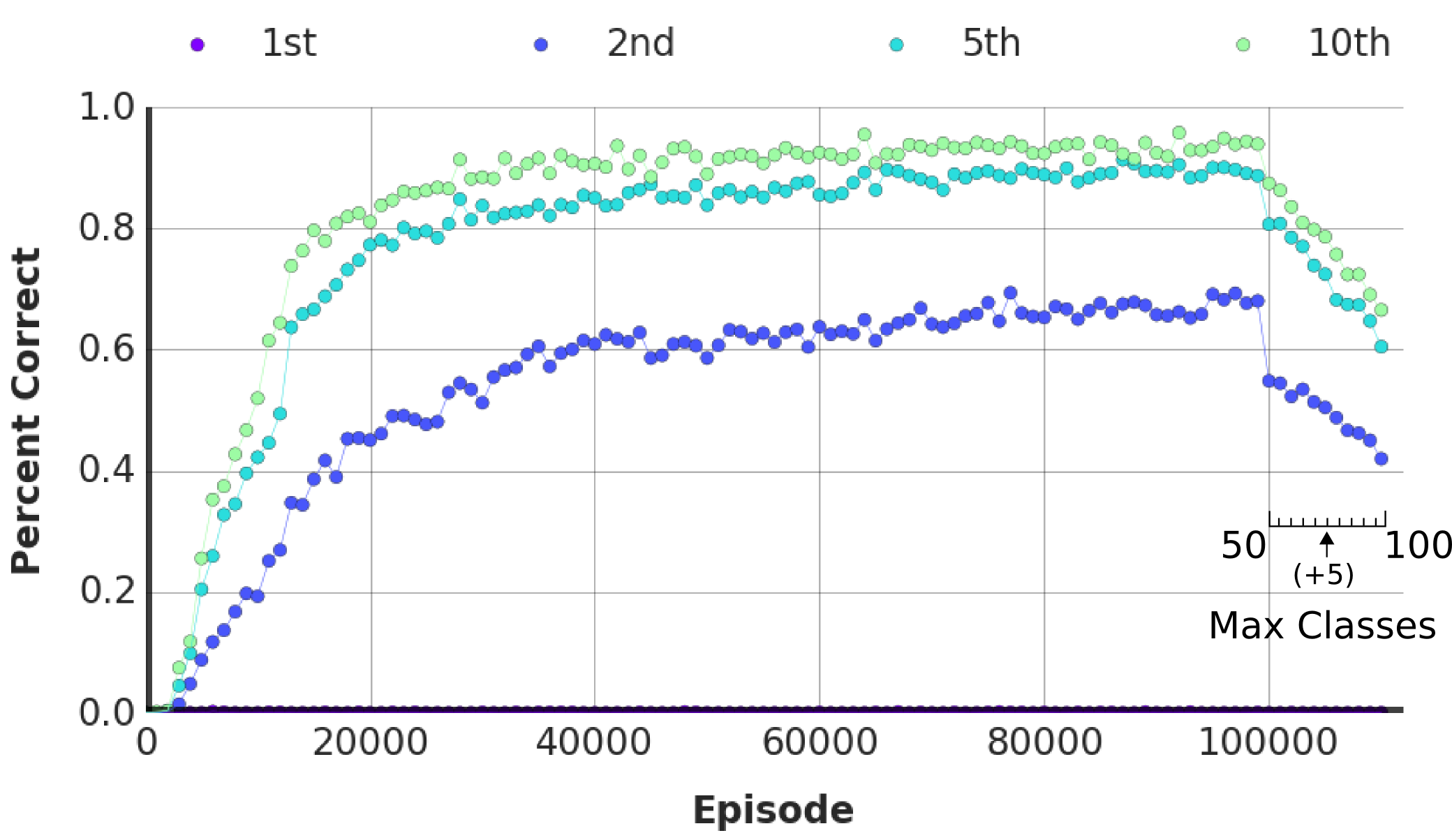}}

\caption{Curriculum classification. The network started with episodes that included up to 15 unique classes, and every 10,000 episodes this maximum was raised by one. Episode lengths were scaled to a value ten times the max number of classes. At the 100,000 episode mark (when the number of classes reached 25) the network was tested on episodes with up to 50 unique classes, which incremented to 100 in steps of five.}%
\label{fig:omniglot-curriculum}%
\end{figure}

Given the successful one-shot classification in episodes with fifteen classes, we employed a curriculum training regime to further scale the classification capabilities of the model. The network was first tasked to classify fifteen classes per episode, and every 10,000 episodes of training thereafter, the maximum number of classes presented per episode incremented by one (Figure \ref{fig:omniglot-curriculum}). The network maintained a high level of accuracy even as the number of classes incremented higher throughout training. After training, at the 100,000 episode mark, the network was tested on episodes with 50 classes. Similar tests continued, increasing the maximum number of classes to 100. The network generally exhibited gradually decaying performance as the number of classes increased towards 100. The training limit of the network seemed to have not been reached, as its performance continued to rise throughout up until the 100,000 episode mark. Assessing the maximum capacity of the network offers an interesting opportunity for future work.

\subsection{Regression}
Since our MANN architecture generated a broad strategy for meta-learning, we reasoned that it would be able to adequately perform regression tasks on never-before-seen functions. To test this, we generated functions using from a GP prior with a fixed set of hyper-parameters and trained our network using unique functions in each episode. Each episode involved the presentation of $x$-values (either 1, 2, or 3-dimensional) along with time-offset function values (i.e., $f(x_{t-1}$)). A successful strategy involves the binding of $x$-values with the appropriate function values and storage of these bindings in the external memory. Since individual $x$-values were only presented once per episode, successful function prediction involved an accurate content-based look-up of proximal information in memory. Thus, unlike in the image-classification scenario, this task demands a broader read from memory: the network must learn to interpolate from previously seen points, which most likely involves a strategy to have a more blended read-out from memory. Such an interpolation strategy in the image classification scenario is less obvious, and probably not necessary. 

\begin{figure}[h]
\centering
\subfloat[][MANN predictions along all x-inputs after 20 samples]{\includegraphics[width=0.98\columnwidth]{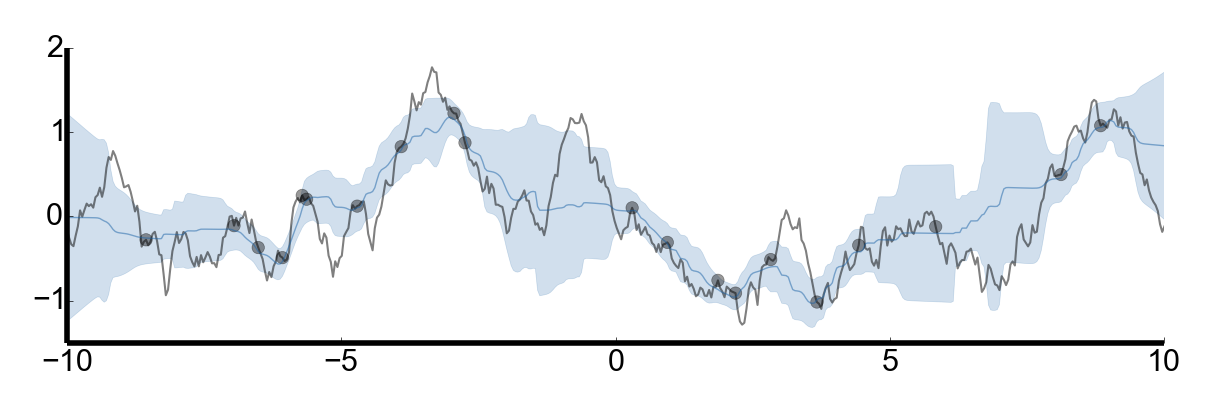}}
\newline
\subfloat[][GP predictions along all x-inputs after 20 samples]{\includegraphics[width=0.98\columnwidth]{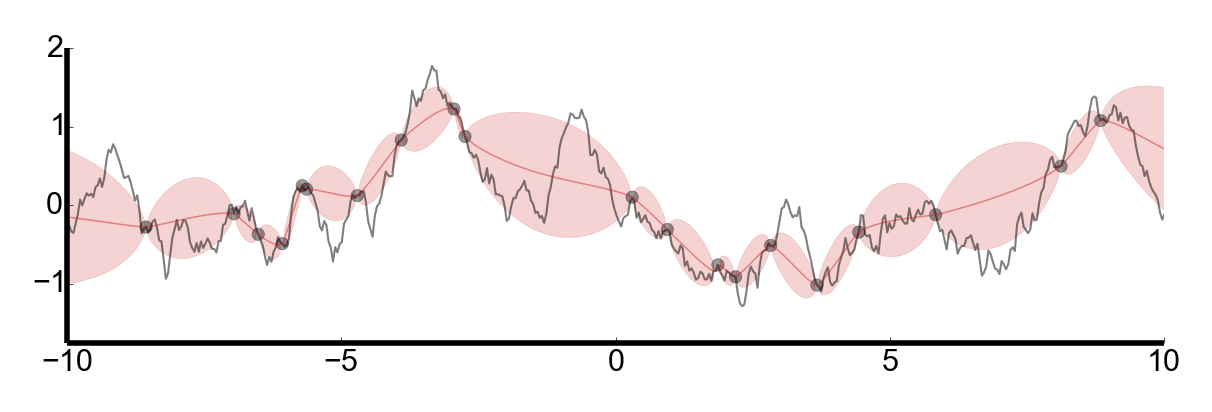}}
\caption{Regression. The network received data samples that were $x$-values for a function sampled from a GP with fixed hyperparameters, and the labels were the associated function values. (a) shows the MANN's predictions for all $x$-values after observing 20 samples, and (b) shows the same for a GP with access to the same hyper-parameters used to generate the function.}%
\label{fig:regression}%
\end{figure}

Network performance was compared to true GP predictions of samples presented in the same order as was seen by the network. Importantly, a GP is able to perform complex queries over all data points (such as covariance matrix inversion) in one step. In contrast, a MANN can only make local updates to its memory, and hence can only approximate such functionality. In our experiments, the GP was initiated with the correct hyper-parameters for the sampled function, giving it an advantage in function prediction. As seen in Figure \ref{fig:regression}, the MANN predictions track the underlying function, with its output variance increasing as it predicts function values that are distance from the values it has already received.

\begin{figure}[h]
\centering
\subfloat[][2D regression log-likelihoods within an episode]{\includegraphics[width=0.98\columnwidth]{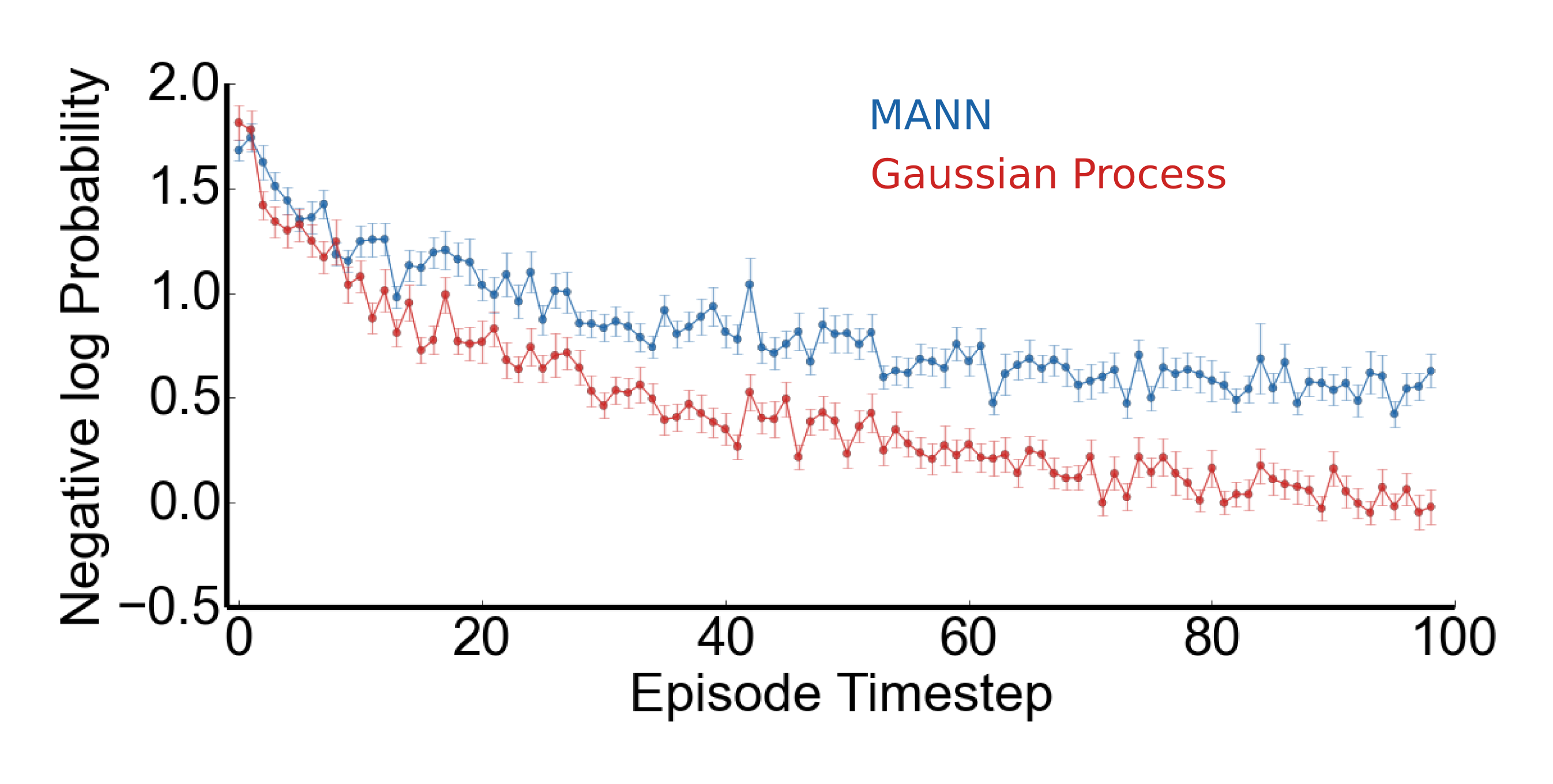}}
\newline
\subfloat[][3D regression log-likelihoods within an episode]{\includegraphics[width=0.98\columnwidth]{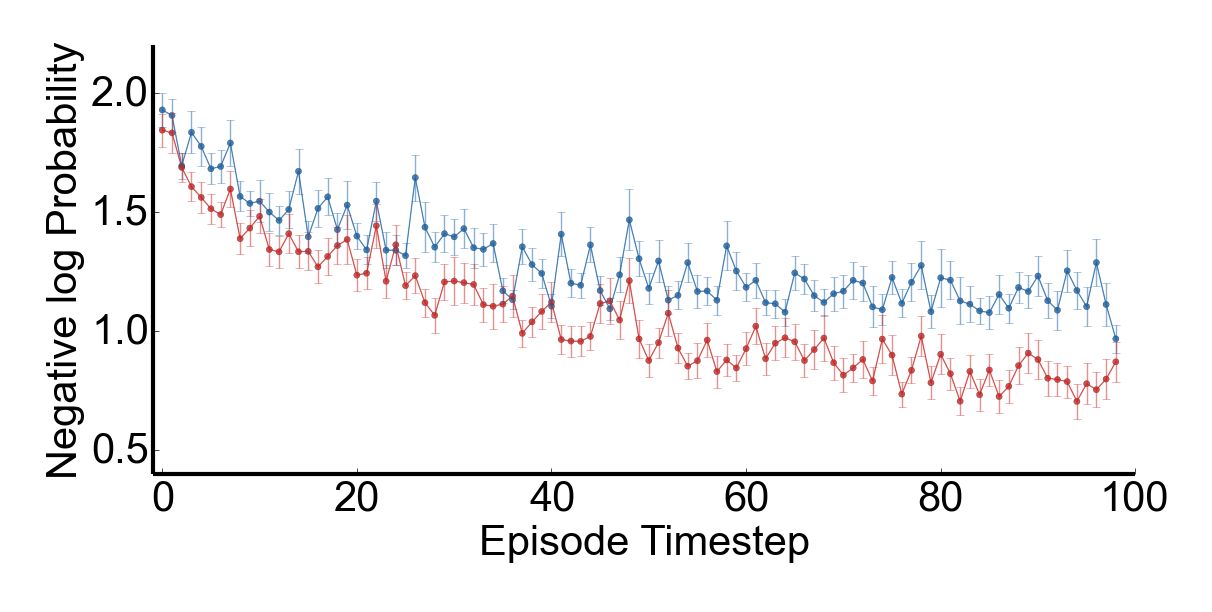}}
\caption{Multi-Dimensional Regression. (a) shows the negative log likelihoods for 2D samples within a single episode, averaged across 100 episodes, while (b) shows the same for 3D samples.}
\label{fig:multiregression}%
\end{figure}

These results were extended to 2-dimensional and 3-dimensional cases  (Fig \ref{fig:multiregression}), with the GP again having access to the correct hyper-parameters for the sampled functions. In both the 2-dimensional and 3-dimensional cases, the log-likelihood predictions of the MANN tracks appreciably well versus the GP, with predictions becoming more accurate as samples are stored in the memory.  

\begin{table*}[t]
\caption{Test-set classification accuracies for various architectures on the Omniglot dataset after 100000 episodes of training, using five-character-long strings as labels. See the supplemental information for an explanation of 1st instance accuracies for the kNN classifier.}
\label{omniglot-string-table}
\vskip 0.15in
\begin{center}
\begin{small}
\begin{sc}
\begin{tabular}{llp{20mm}|p{8mm}p{8mm}p{8mm}p{8mm}p{8mm}p{8mm}}
\hline
& & &  \multicolumn{6}{c}{Instance (\% Correct)} \\
Model & Controller &  \# of Classes & 1\textsuperscript{st} & 2\textsuperscript{nd} & 3\textsuperscript{rd} & 4\textsuperscript{th} & 5\textsuperscript{th} & 10\textsuperscript{th} \\
\hline
\abovespace
kNN (raw pixels) &   -- & 5  & 4.0 & 36.7 & 41.9 & 45.7 & 48.1 & 57.0 \\
kNN (deep features) & --          & 5  & 4.0 & 51.9 & 61.0 & 66.3 & 69.3 & 77.5 \\
Feedforward &   --           & 5  & 0.0     & 0.2             & 0.0           & 0.2           & 0.0         & 0.0\\
LSTM        &   --           & 5  & 0.0     & 9.0             & 14.2          & 16.9          & 21.8        & 25.5\\
MANN        &   Feedforward  & 5  & 0.0     & 8.0             & 16.2          & 25.2          & 30.9          & 46.8  \\
MANN        &   LSTM         & 5  & 0.0     & \textbf{69.5}   & \textbf{80.4} & \textbf{87.9} & \textbf{88.4} & \textbf{93.1}  \\
\hline
\abovespace
kNN (raw pixels) &   --           & 15 & 0.5 & 18.7 & 23.3 & 26.5 & 29.1 & 37.0 \\
kNN (deep features) &   --           & 15 & 0.4 & 32.7 & 41.2 & 47.1 & 50.6 & 60.0  \\
Feedforward &   --           & 15 & 0.0     & 0.1           & 0.0           & 0.0           & 0.0             & 0.0  \\
LSTM        &   --           & 15 & 0.0     & 2.2           & 2.9           & 4.3           & 5.6             & 12.7  \\
MANN (LRUA) &   Feedforward  & 15 & 0.1     & 12.8          & 22.3          & 28.8          & 32.2          & 43.4  \\
MANN (LRUA) &   LSTM         & 15 & 0.1     & \textbf{62.6} & \textbf{79.3} & \textbf{86.6} & \textbf{88.7} & \textbf{95.3}  \\
MANN (NTM)&LSTM         & 15 & 0.0     & 35.4            & 61.2          & 71.7          & 77.7          & 88.4 \\
\hline
\end{tabular}
\end{sc}
\end{small}
\end{center}
\vskip -0.1in
\end{table*}

\section{Discussion \& Future Work}
Many important learning problems demand an ability to draw valid inferences from small amounts of data, rapidly and knowledgeably adjusting to new information. Such problems pose a particular challenge for deep learning, which typically relies on slow, incremental parameter changes. We investigated an approach to this problem based on the idea of meta-learning. Here, gradual, incremental learning encodes background knowledge that spans tasks, while a more flexible memory resource binds information particular to newly encountered tasks. Our central contribution is to demonstrate the special utility of a particular class of MANNs for meta-learning. These are deep-learning architectures containing a dedicated, addressable memory resource that is structurally independent from the mechanisms that implement process control. The MANN examined here was found to display performance superior to a LSTM in two meta-learning tasks, performing well in classification and regression tasks when only sparse training data was available. 

A critical aspect of the tasks studied is that they cannot be performed based solely on rote memory. New information must be flexibly stored and accessed, with correct performance demanding more than just accurate retrieval. Specifically, it requires that inferences be drawn from new data based on longer-term experience, a faculty sometimes referred as ``inductive transfer."  MANNs are well-suited to meet these dual challenges, given their combination of flexible memory storage with the rich capacity of deep architectures for representation learning. 

Meta-learning is recognized as a core ingredient of human intelligence, and an essential test domain for evaluating models of human cognition. Given recent successes in modeling human skills with deep networks, it seems worthwhile to ask whether MANNs embody a promising hypothesis concerning the mechanisms underlying human meta-learning. In informal comparisons against human subjects, the MANN employed in this paper displayed superior performance, even at set-sizes that would not be expected to overtax human working memory capacity. However, when memory is not cleared between tasks, the MANN suffers from proactive interference, as seen in many studies of human memory and inference \cite{underwood1957interference}. These preliminary observations suggest that MANNs may provide a useful heuristic model for further investigation into the computational basis of human meta-learning. 

The work we presented leaves several clear openings for next-stage development. First, our experiments employed a new procedure for writing to memory that was \textit{prima facie} well suited to the tasks studied. It would be interesting to consider whether meta-learning can itself discover optimal memory-addressing procedures. Second, although we tested MANNs in settings where task parameters changed across episodes, the tasks studied contained a high degree of shared high-level structure. Training on a wider range of tasks would seem likely to reintroduce standard challenges associated with continual learning, including the risk of catastrophic interference. Finally, it may be of interest to examine MANN performance in meta-learning tasks requiring active learning, where observations must be actively selected. 

\clearpage

\section{Acknowledgements}
The authors would like to thank Ivo Danihelka and Greg Wayne for helpful discussions and prior work on the NTM and LRU Access architectures, as well as Yori Zwols, and many others at Google DeepMind for reviewing the manuscript.

\clearpage

\bibliography{paper}
\bibliographystyle{icml2016_arxiv}

\clearpage

\section*{Supplementary Information}
\subsection{Additional model details}
Our model is a variant of a Neural Turing Machine (NTM) from Graves et al. It consists of a number of differentiable components: a controller, read and write heads, an external memory, and an output distribution. The controller receives input data (see section \ref{sec:datasup}) directly, and also provides an input to the output distribution. Each of these components will be addressed in turn.

\begin{figure}[h]
\centering
\subfloat[][]{\includegraphics[width=0.7\columnwidth]{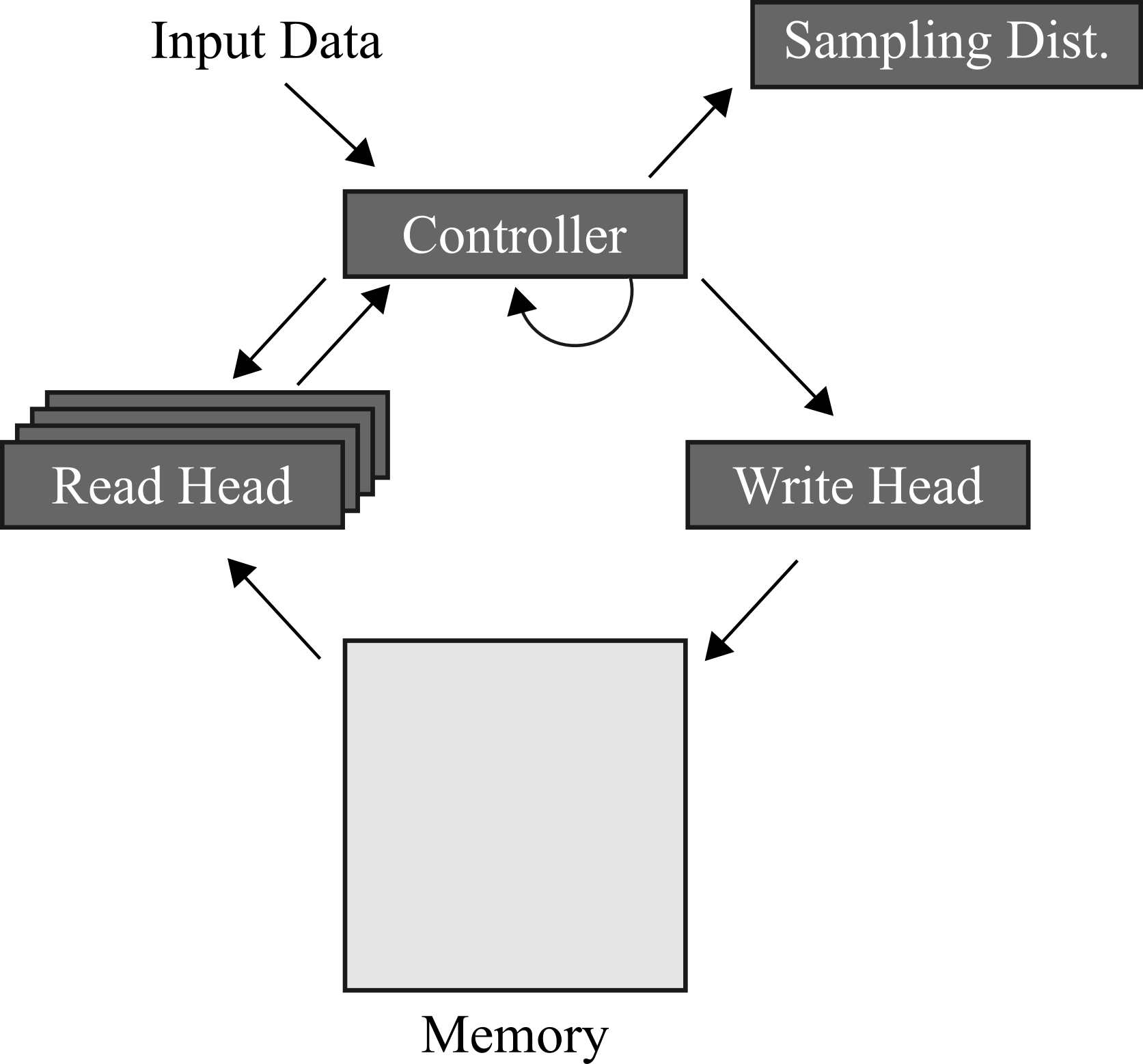}}
\caption{MANN Architecture.}
\label{fig:ntm}%
\end{figure}

The controllers in our experiments are feed-forward networks or Long Short-Term Memories (LSTMs). For the best performing networks, the controller is a LSTM with 200 hidden units. The controller receives some concatenated input $(\mathbf{x}_t, \mathbf{y}_{t-1})$ (see section \ref{sec:datasup} for details) and updates its state according to:
\begin{align}
    \mathbf{\hat{g}}^f, \mathbf{\hat{g}}^i, \mathbf{\hat{g}}^o, \mathbf{\hat{u}} &= \mathbf{W}^{xh}(\mathbf{x}_t, \mathbf{y}_{t-1}) + \mathbf{W}^{hh}\mathbf{h}_{t-1} + \mathbf{b}^h, \\
    \mathbf{g}^f &= \sigma(\mathbf{\hat{g}}^f), \\
    \mathbf{g}^i &= \sigma(\mathbf{\hat{g}}^i), \\
    \mathbf{g}^o &= \sigma(\mathbf{\hat{g}}^o), \\
    \mathbf{u} &= \text{tanh}(\mathbf{\hat{u}}), \\
    \mathbf{c}_t &= \mathbf{g}^f \odot \mathbf{c}_{t-1} + \mathbf{g}^i \odot \mathbf{u}, \\
    \mathbf{h}_t &= \mathbf{g}^o \odot \text{tanh}(\mathbf{c}_t), \\
    \mathbf{o}_t &= (\mathbf{h}_t, \mathbf{r}_t)
\end{align}
where $\mathbf{\hat{g}}^f$, $\mathbf{\hat{g}}^o$, and $\mathbf{\hat{g}}^i$ are the forget gates, output gates, and input gates, respectively, $\mathbf{b}^h$ are the hidden state biases, $\mathbf{c}_t$ is the cell state, $\mathbf{h}_t$ is the hidden state, $\mathbf{r}_t$ is the vector read from memory, $\mathbf{o}_t$ is the concatenated output of the controller, $\odot$ represents element-wise multiplication, and $(\cdot,\cdot)$ represents vector concatenation. $\mathbf{W}^{xh}$ are the weights from the input $(\mathbf{x}_t, \mathbf{y}_{t-1})$ to the hidden state, and $\mathbf{W}^{hh}$ are the weights between hidden states connected through time. The read vector $\mathbf{r}_t$ is computed using content-based addressing using a cosine distance measure, as described in the main text, and is repeated below for self completion.

The network has an external memory module, $\mathbf{M}_t$, that is both read from and written to. The rows of $\mathbf{M}_t$ serve as memory `slots', with the row vectors themselves constituting individual memories. For reading, the controller cell state serves as a query for $\mathbf{M}_t$. First, a cosine distance measure is computed for the query key vector (here notated as $\mathbf{k}_t$) and each individual row in memory:
\begin{align}
    K\big(\mathbf{k}_t, \mathbf{M}_{t}(i)\big) = \frac{\mathbf{k}_t \cdot \mathbf{M}_{t}(i)}{\parallel\mathbf{k}_t\parallel\parallel\mathbf{M}_{t}(i)\parallel},
\end{align}

Next, these similarity measures are used to produce a read-weight vector $\mathbf{w}_t^r$, with elements computed according to a softmax:
\begin{align}
    w_t^{r}(i) \leftarrow \frac{\text{exp}\big(K\big(\mathbf{k}_t, \mathbf{M}_{t}(i)\big)\big)}{\sum_j \text{exp} \big( K\big(\mathbf{k}_t, \mathbf{M}_{t}(j)\big)\big)}.
\end{align}

A memory, $\mathbf{r}_t$, is then retrieved using these read-weights:
\begin{align}
    \mathbf{r}_t \leftarrow \sum_i {w}_t^{r}(i)\mathbf{M}_{t}(i).
\end{align}

Finally, $\mathbf{r}_t$ is concatenated with the controller hidden state, $\mathbf{h}_t$, to produce the network's output $\mathbf{o}_t$ (see equation (16)). The number of reads from memory is a free parameter, and both one and four reads were experimented with. Four reads was ultimately chosen for the reported experimental results. Multiple reads is implemented as additional concatenation to the output vector, rather than any sort of combination or interpolation. 

To write to memory, we implemented a new content-based access module called Least Recently Used Access (LRUA). LRUA writes to either the most recently read location, or the least recently used location, so as to preserve recent, and hence potentially useful memories, or to update recently encoded information. Usage weights $\mathbf{w}_t^u$ are computed each time-step to keep track of the locations most recently read or written to:
\begin{align}
    \mathbf{w}_t^u \leftarrow \gamma \mathbf{w}_{t-1}^u + \mathbf{w}_t^r + \mathbf{w}_t^w,
\end{align}

where $\gamma$ is a decay parameter. The \textit{least-used} weights, $\mathbf{w}_t^{lu}$, for a given time-step can then be computed using $\mathbf{w}_{t}^u$. First, we introduce the notation $m(\mathbf{v}, n)$ to denote the $n^{th}$ smallest element of the vector $\mathbf{v}$. Elements of $\mathbf{w}_{t}^{lu}$ are set accordingly:
\begin{align}
    w_{t}^{lu}(i) = 
    \left\{
	\begin{array}{ll}
		0  & \mbox{if } w_{t}^{u}(i) > m(\mathbf{w}_{t}^{u}, n) \\
		1  & \mbox{if } w_{t}^{u}(i) \le m(\mathbf{w}_{t}^{u}, n)
	\end{array}
\right. ,
\end{align}
where $n$ is set to equal the number of reads to memory.

To obtain the write weights $\mathbf{w}_t^w$, a learnable sigmoid gate parameter is used to compute a convex combination of the previous read weights and previous least-used weights:
\begin{align}
    \mathbf{w}_t^w \leftarrow \sigma(\alpha) \mathbf{w}_{t-1}^r + (1-\sigma(\alpha))\mathbf{w}_{t-1}^{lu},
\end{align}

where $\alpha$ is a dynamic scalar gate parameter to interpolate between the weights. Prior to writing to memory, the least used memory location is computed from $\mathbf{w}_{t-1}^u$ and is set to zero. Writing to memory then occurs in accordance with the computed vector of write weights:
\begin{align}
    \mathbf{M}_t(i) \leftarrow \mathbf{M}_{t-1}(i) + w_t^{w}(i) \mathbf{k}_{t}, \forall i
\end{align}

\subsection{Output distribution}
The controller's output, $\mathbf{o}_t$, is propagated to an output distribution. For classification tasks using one-hot labels, the controller output is first passed through a linear layer with an output size equal to the number of classes to be classified per episode. This linear layer output is then passed as input to the output distribution. For one-hot classification, the output distribution is a categorical distribution, implemented as a softmax function. The categorical distribution produces a vector of class probabilities, $\mathbf{p}_t$, with elements:
\begin{align}
    p_t(i) = \frac{\text{exp}(\mathbf{W}^{op}(i)\mathbf{o}_t)}{\sum_{j}\text{exp}(\mathbf{W}^{on}(j)\mathbf{o}_t)},
\end{align}

where $\mathbf{W}^{op}$ are the weights from the controller output to the linear layer output.

For classification using string labels, the linear output size is kept at 25. This allows for the output to be split into five equal parts each of size five. Each of these parts is then sent to an independent categorical distribution that computes probabilities across its five inputs. Thus, each of these categorical distributions independently predicts a `letter,' and these letters are then concatenated to produce the five-character-long string label that serves as the network's class prediction (see figure \ref{fig:sequence}). 

A similar implementation is used for regression tasks. The linear output from the controller outputs two values: $\mu$ and $\sigma$, which are passed to a Gaussian distribution sampler as predicted mean and variance values. The Gaussian sampling distribution then computes probabilities for the target value $y_t$ using these values.

\subsection{Learning}
For one-hot label classification, given the probabilities output by the network, $\mathbf{p}_t$, the network minimizes the episode loss of the input sequence:
\begin{align}
    \mathcal{L}(\theta) = - \sum_t \mathbf{y}_t^{T} \log \mathbf{p}_t,
\end{align}
where $\mathbf{y}_t$ is the target one-hot or string label at time $t$ (note: for a given one-hot class-label vector $\mathbf{y}_t$, only one element assumes the value 1, and for a string-label vector, five elements assume the value 1, one per five-element `chunk').

For string label classification, the loss is similar:
\begin{align}
    \mathcal{L}(\theta) = - \sum_t \sum_c \mathbf{y}_t^{T}(c) \log \mathbf{p}_t(c).
\end{align}
Here, the $(c)$ indexes a five-element long `chunk' of the vector label, of which there are a total of five.

For regression, the network's output distribution is a Gaussian, and as such receives two-values from the controller output's linear layer at each time-step: predictive $\mu$ and $\sigma$ values, which parameterize the output distribution. Thus, the network minimizes the negative log-probabilities as determined by the Gaussian output distribution given these parameters and the true target $y_t$.

\section{Classification input data}
\label{sec:datasup}
\begin{figure}[h]
\centering
\subfloat[][String label encoded as five-hot vector]{\includegraphics[width=0.9\columnwidth]{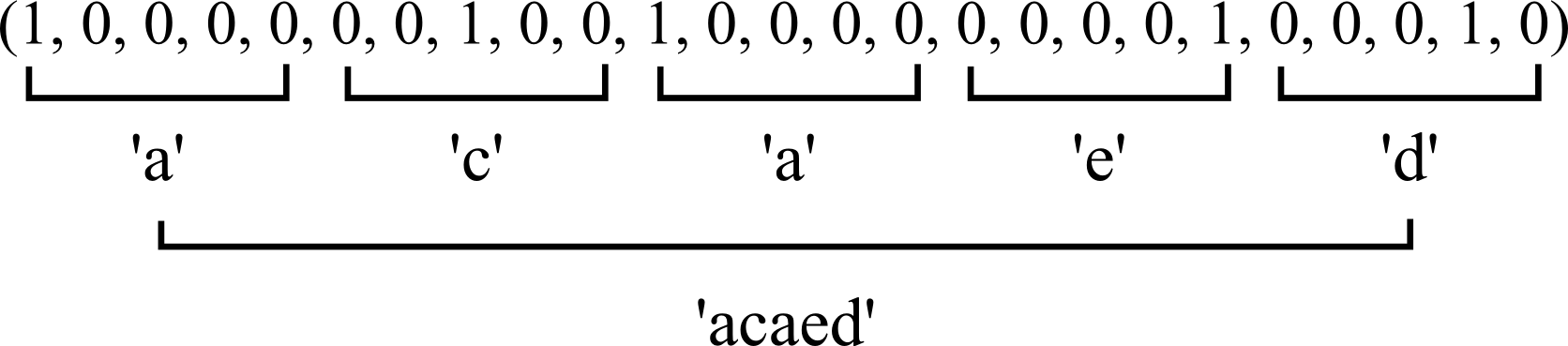}}
\qquad
\subfloat[][Input Sequence]{\includegraphics[width=0.9\columnwidth]{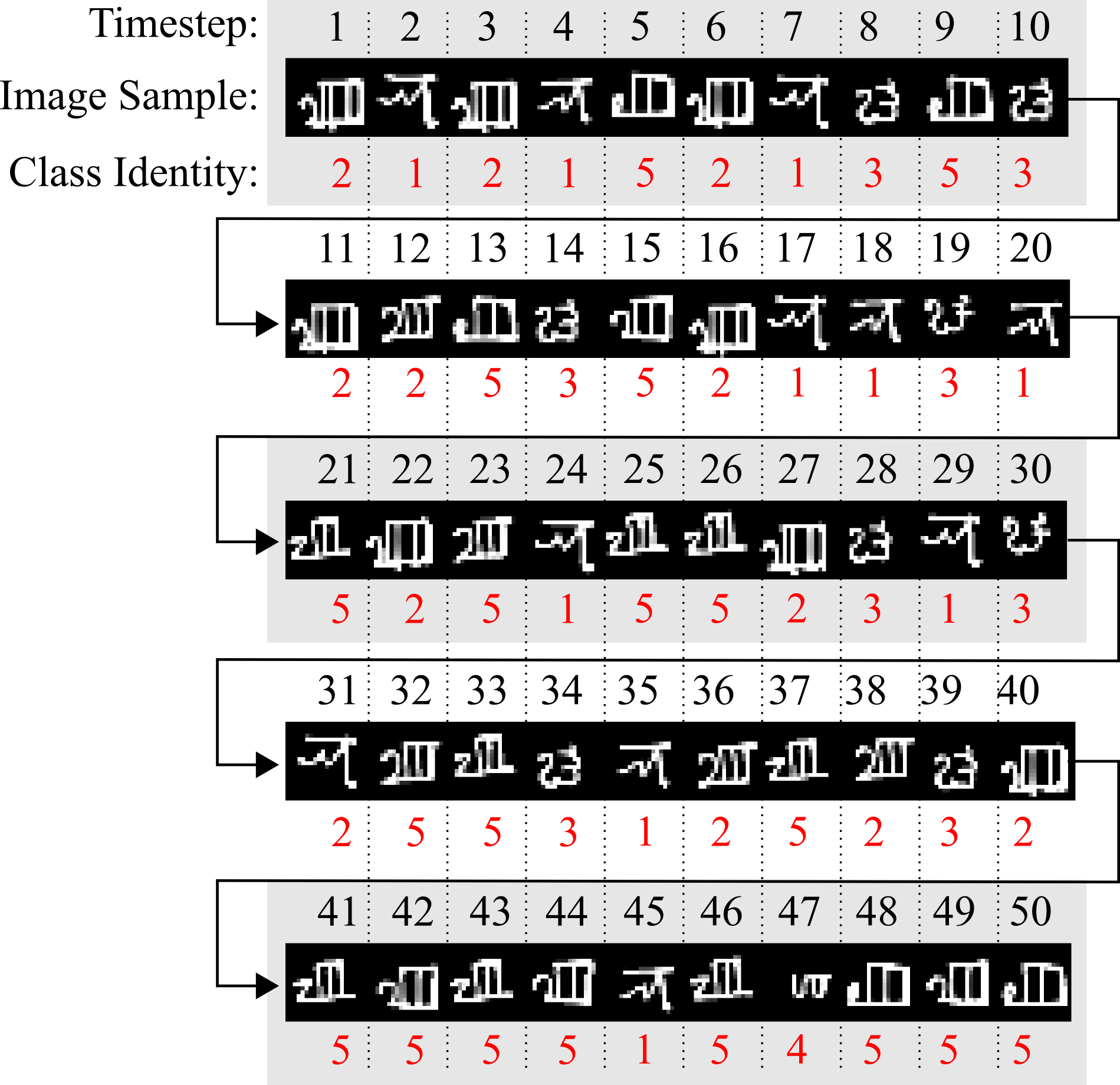}}
\caption{Example string label and input sequence.}
\label{fig:sequence}%
\end{figure}

Input sequences consist of flattened, pixel-level representations of images $\mathbf{x}_t$ and time-offset labels $\mathbf{y}_{t-1}$ (see figure \ref{fig:sequence} for an example sequence of images and class identities for an episode of length 50, with five unique classes). First, $N$ unique classes are sampled from the Omniglot dataset, where $N$ is the maximum number of unique classes per episode. $N$ assumes a value of either 5, 10, or 15, which is indicated in the experiment description or table of results in the main text. Samples from the Omniglot source set are pulled, and are kept if they are members of the set of $n$ unique classes for that given episode, and discarded otherwise. $10N$ samples are kept, and constitute the image data for the episode. And so, in this setup, the number of samples per unique class are not necessarily equal, and some classes may not have any representative samples. Omniglot images are augmented by applying a random rotation uniformly sampled between $-\frac{\pi}{16}$ and $\frac{\pi}{16}$, and by applying a random translation in the x- and y- dimensions uniformly sampled between -10 and 10 pixels. The images are then downscaled to $20\text{x}20$. A larger class-dependent rotation is then applied, wherein each sample from a particular class is rotated by either $0$, $\frac{\pi}{2}$, $\pi$, or $\frac{3\pi}{2}$ (note: this class-specific rotation is randomized each episode, so a given class may experience different rotations from episode-to-episode). The image is then flattened into a vector, concatenated with a randomly chosen, episode-specific label, and fed as input to the network controller.

Class labels are randomly chosen for each class from episode-to-episode. For one-hot label experiments, labels are of size $N$, where $N$ is the maximum number of unique classes that can appear in a given episode. 

\section{Task}
Either 5, 10, or 15 unique classes are chosen per episode. Episode lengths are ten times the number of unique classes (i.e., 50, 100, or 150 respectively), unless explicitly mentioned otherwise. Training occurs for 100 000 episodes. At the 100 000 episode mark, the task continues; however, data are pulled from a disjoint test set (i.e., samples from classes 1201-1623 in the omniglot dataset), and weight updates are ceased. This is deemed the ``test phase."

For curriculum training, the maximum number of unique classes per episode increments by 1 every 10 000 training episodes. Accordingly, the episode length increases to 10 times this new maximum.

\section{Parameters}
\subsubsection{Optimization}
Rmsprop was used with a learning rate of $1\text{e}^{-4}$ and max learning rate of $5\text{e}^{-1}$, decay of 0.95 and momentum 0.9. 

\subsubsection{Free parameter grid search}
A grid search was performed over number of parameters, with the values used shown in parentheses: memory slots (128), memory size (40), controller size (200 hidden units for a LSTM), learning rate ($1\text{e}^{-4}$), and number of reads from memory (4). Other free parameters were left constant: usage decay of the write weights (0.99), minibatch size (16),

\subsection{Comparisons and controls evaluation metrics}
\subsubsection{Human comparison}
For the human comparison task, participants perform the exact same experiment as the network: they observe sequences of images and time-offset labels (sequence length = 50, number of unique classes = 5), and are challenged to predict the class identity for the current input image by inputting a single digit on a keypad. However, participants view class labels the integers 1 through 5, rather than one-hot vectors or strings. There is no time limit for their choice. Participants are made aware of the goals of the task prior to starting, and they perform a single, non-scored trial run prior to their scored trials. Nine participants each performed two scored trials.

\subsubsection{kNN}
When no data is available (i.e., at the start of training), the kNN classifier randomly returns a single class as its prediction. So, for the first data point, the probability that the prediction is correct is $\frac{1}{N}$ where $N$ is number of unique classes in a given episode. Thereafter, it predicts a class from classes that it has observed. So, all instances of samples that are not members of the first observed class cannot be correctly classified until at least one instance is passed to the classifier. Since statistics are averaged across classes, first instance accuracy becomes $\frac{1}{N}(\frac{1}{N} + 0) = \frac{1}{N^2}$, which is 4\% and 0.4\% for 5 and 15 classes per episode, respectively. 

\end{document}